\documentclass[conference,compsoc]{IEEEtran}
\usepackage{graphicx}
\usepackage{subfigure}
\usepackage{float}
\usepackage{verbatim}
\usepackage{multirow}

\ifCLASSOPTIONcompsoc
  \usepackage[nocompress]{cite}
\else
  \usepackage{cite}
\fi
\ifCLASSINFOpdf
\else
\fi

\usepackage{xcolor}
\usepackage{xspace}

\makeatletter
\newcommand{\linebreakand}{%
  \end{@IEEEauthorhalign}
  \hfill\mbox{}\par
  \mbox{}\hfill\begin{@IEEEauthorhalign}
}
\makeatother

\begin{document}
%

%
\title{Utilizing Ensemble Learning for Performance and Power Modeling and Improvement of Parallel Cancer Deep Learning CANDLE Benchmarks}
%


\author{\IEEEauthorblockN{Xingfu Wu}
\IEEEauthorblockA{Argonne National Laboratory\\ The University of Chicago, USA \\
Email: xingfu.wu@anl.gov}
\and
\and
\IEEEauthorblockN{Valerie Taylor}
\IEEEauthorblockA{Argonne National Laboratory\\ The University of Chicago, USA \\
Email: vtaylor@anl.gov}
}


\maketitle

\thispagestyle{plain}
\pagestyle{plain}

\begin{abstract}

Machine learning  (ML) continues to grow in importance across nearly all domains and is a natural tool in modeling to learn from data. Often a tradeoff exists between a model's ability to minimize bias and variance. In this paper, we utilize ensemble learning to combine linear, nonlinear, and tree-/rule-based ML methods to cope with the bias-variance tradeoff and result in more accurate models. Hardware performance counter values are correlated with properties of applications that impact performance and power on the underlying system. We use the datasets collected for two parallel cancer deep learning CANDLE benchmarks, NT3 (weak scaling) and P1B2 (strong scaling), to build performance and power models based on hardware performance counters using single-object and multiple-objects ensemble learning to identify the most important counters for improvement. Based on the insights from these models, we improve the performance and energy of P1B2 and NT3 by optimizing the deep learning environments  TensorFlow, Keras, Horovod, and Python under the huge page size of 8 MB on the Cray XC40 Theta at Argonne National Laboratory. Experimental results show that ensemble learning not only produces more accurate models but also provides more robust performance counter ranking. We achieve up to 61.15\% performance improvement and up to 62.58\% energy saving for P1B2 and up to 55.81\% performance improvement and up to 52.60\% energy saving for NT3 on up to 24,576 cores.

\end{abstract}


\section{Introduction}

Energy-efficient scientific applications require insight into how high-performance computing (HPC) system components impact the applications' power and performance. This insight can result from the development of performance and power models. Currently, HPC systems, especially petaflops supercomputers, consume a huge amount of power; and exascale HPC systems will be similarly constrained. Therefore, monitoring the power consumption of an HPC system is important for power management. 
Since direct online power measurement at high frequencies is impractical, hardware performance counters have been widely used as effective proxies to estimate power consumption \cite{SB08, BJ12, RA13}. Hardware performance counter values are correlated with properties of applications that impact performance and power on the underlying system. In this paper, we use ensemble machine learning, which combines several machine learning models to create a new ensemble model \cite{MZ16, FL12} for modeling performance and power based on performance counters and rank performance counters, with the aim of  identifying the most important performance counters for application improvement.

Much of the previous work on power modeling and estimation is based on performance counters \cite{IM03, CM05, CD06, SB08, LP10}, 
\cite{CX10, NM10, BJ12, LW12, RA13, SS13, LT14, TL14, WT16, WT16b, ZJ17, GL18}. These counters are used to monitor system components such as CPU, GPU, memory, disk, and I/O. The values of these performance counters are then correlated with the power consumed by each system component to derive a power model for each system component. Many of these approaches use a small set of performance counters (13 counters or less) for power modeling. In our previous work \cite{LT14, WT16}, 
we developed models of runtime and power based on 40  performance counters and found that the performance counters used for four different models (runtime, node power, CPU power, and memory power) were not the same. For instance, with six scientific applications we found that a total of 37 different performance counters were used for the models. 
Hence, in this paper we use ensemble machine learning to enhance our modeling methods. 

Machine learning  (ML) continues to grow in importance across nearly all domains and is a natural tool in modeling to learn from data. It is the process of developing a model in a way that we can understand and quantify the model's prediction accuracy on future, yet-to-be-seen data \cite{KJ13}. 
A variety of ML models have been fit to different datasets in different languages. For instance, the machine learning caret package in R \cite{1} provides 238 machine learning methods in R, and the scikit-learn \cite{2} package provides many machine learning methods in Python. ML methods have their own way of learning the relationship between the predictors and the target object. 
Which method, then, should be selected for the final model? 

 Generally, prediction errors can be decomposed into two important subcomponents: error due to bias and error due to variance \cite{1}. Error due to bias is the difference between the expected (or average) prediction of a model and the actual value. It measures how far off the model's predictions are from the actual value, which provides a sense of how well the model can conform to the underlying structure of the data. On the other hand, error due to variance is defined as the variability of a model prediction for a given data point.  Generally,  more complex models such as k-nearest neighbor, decision trees, and gradient boosting machines can have very high variance, which leads to overfitting. Simple models such as linear models, which are classical examples of high-bias models, tend not to overfit but to underfit because they are less flexible and rarely capture nonlinear relationships. Understanding how different sources of error lead to bias and variance helps us improve the data-fitting process, resulting in more accurate models. Often a tradeoff must be made between a model's ability to minimize bias and variance.

Ensemble learning combines several ML models to create a new ensemble model \cite{MZ16, FL12}. It is a powerful technique in machine learning, often outperforming other methods with respect to accuracy. However, this comes at the cost of increased algorithmic and model complexity. Should the ensemble model be selected for the final model? Should the ensemble learning provide the robust feature importance? When we consider multiple objects such as runtime, node power, CPU power, and memory power for improvement, how do we utilize multiple-object ensemble learning? In this paper, we address these issues and use ensemble learning to combine linear, nonlinear, and tree-/rule-based ML methods to cope with the bias-variance tradeoff and result in more accurate models.

The CANDLE project \cite{8, 21, WT19} focuses on building a single scalable deep neural network that can address three cancer challenge problems: (1) the RAS pathway problem of understanding the molecular basis of key protein interactions in the RAS/RAF pathway presented in 30\% of cancers using unsupervised learning; (2) the drug response problem of developing predictive models for drug response to optimize preclinical drug screening and drive precision-medicine-based treatments for cancer patients using supervised learning; and (3) the treatment strategy problem of automating the analysis and extraction of information from millions of cancer patient records to determine optimal cancer treatment strategies using semi-supervised learning. The CANDLE benchmarks \cite{21} implement deep learning architectures that are relevant to these three cancer problems. In our previous work \cite{WT19}, we discussed our parallel methodology and implemented and analyzed the Horovod CANDLE benchmarks (NT3, P1B1, P1B2, and P1B3), focusing on their scalability, performance, and power characteristics with different batch sizes, learning rates, and epochs on both Summit \cite{SUMM} with GPUs at Oak Ridge National Laboratory and Theta \cite{THET} with CPUs at Argonne National Laboratory. We identified the data-loading bottlenecks and then improved the performance and energy for better scalability by loading data in chunks without memory limitation. We achieved up to 78.25\% in performance improvement and up to 78\% in energy saving under strong scaling on up to 384 GPUs and up to 79.5\% in performance improvement and up to 77.11\% in energy saving under weak scaling on up to 3,072 GPUs on Summit. We also achieved up to 45.22\% performance improvement and up to 41.78\% in energy saving under strong scaling on up to 384 nodes on Theta. However,  these benchmarks still ran much slower than what we expected on Theta. This result motivated us to identify opportunities to further improve their performance and energy on Theta. 

In this paper, we use the datasets collected for the parallel cancer deep learning CANDLE benchmarks NT3 (weak scaling) and P1B2 (strong scaling) to address these issues in performance and power modeling and to achieve improvement using single-object and multiple-object ensemble learning \cite{MZ16, MP16}, which is a combination of several machine learning methods from the R package caret. We focus on three types of machine learning: linear, nonlinear, and tree-/rule-based regressions. We analyze how a single ML method performs, then investigate how ensemble learning for 15 ML methods performs and compare the results and evaluate models with built-in feature selection. Further, we use unsupervised feature selection methods to confirm that ensemble learning provides more robust model and feature ranking. Then, based on the insights from these models, we improve the performance and energy of P1B2 and NT3 on Theta.

The remainder of this paper is organized as follows. Section 2 presents the modeling and improvement framework using ensemble learning. Section 3 briefly describes the parallel cancer deep learning CANDLE benchmarks NT3 and P1B2. Section 4 discusses performance and power modeling using single-object ensemble learning and investigates the performance counter ranking under different ML methods. Section 5 discusses performance and power modeling using multiple-objects ensemble learning. Section 6 presents performance and energy improvement and   the experimental results. Section 7 summarizes our conclusions and briefly discusses possible future work.

\section{Modeling and Improvement Framework Using Ensemble Learning }

In this section, we propose a modeling and improvement framework using ensemble learning. We  discuss how to use ensemble machine learning methods to model performance and power, and we identify the most important performance counters that affect the application performance and power. 

Because our problem is a regression problem, we focus on three types of machine learning with a total of 15 methods: linear, nonlinear, and tree-/rule-based regressions with built-in measurements of variable importance from the caret package in R \cite{1}. For instance, multivariate adaptive regression spline  and many tree-based models monitor the increase in performance that occurs when adding each variable to the model \cite{KJ13}, and linear regression and logistic regression use quantifications based on the model coefficients or statistical measurements (such as t-statistics). 

For linear regression, we use five methods: Lasso and elastic-net regularized generalized linear models (glmnet) \cite{GLM}, partial least squares (pls) \cite{KJ13}, ridge regression \cite{ZH18}, principal component regression (pcr) \cite{KJ13}, and elastic net regression (enet) \cite{FH10}. 
For nonlinear regression, we use five methods: k-nearest neighbors (knn) \cite{KJ13}, support vector machine with a linear kernel \cite{KSH}, multivariate adaptive regression spline \cite{MS19}, Gaussian process with radial basis function (gaussprRadial) \cite{KSH}, and bayesglm \cite{KJ13}. For tree-/rule-based regression, we use five methods: random forests (rf) \cite{LW18}, cubist \cite{KW20},  conditional inference tree (ctree) \cite{HH20}, eXtreme Gradient Boosting (xgbTree) \cite{CH19}, and bagged (bootstrap aggregated) CART (Classification \& Regression Trees) (treebag) \cite{KJ13}. 

To collect these ML models as an ensemble, we use caretEnsemble \cite{MZ16}, which is a package for making ensembles of the caret models from R caret \cite{1}. The caretEnsemble package has three primary functions: caretList(), caretEnsemble(), and caretStack(). caretList is a function for fitting many different caret models to the same dataset with the same resampling parameters. It returns a list of caret objects that can be passed to caretEnsemble or caretStack. It has almost the same arguments as train() from the R caret. caretEnsemble uses a generalized linear model (glm) to create a simple linear blend of models. It has two arguments that can be used to specify which models to fit: methodList and tuneList. methodList is a simple character vector of methods that will be fit with the default train parameters, while tuneList can be used to customize the call to each component model. varImp() \cite{1} is used to extract the variable importance from each member of the ensemble, as well as the final ensemble model. caretStack uses a caret model to combine the outputs from several component caret models. It allows to move beyond simple blends of models to use metamodels to create ensemble collections of predictive models. However, it does not support varImp(). In this work, we use caretEnsemble to make an ensemble of the 15 ML methods to build the model and rank the performance counters.

\begin{figure}
\center
 \includegraphics[width=.48\textwidth]{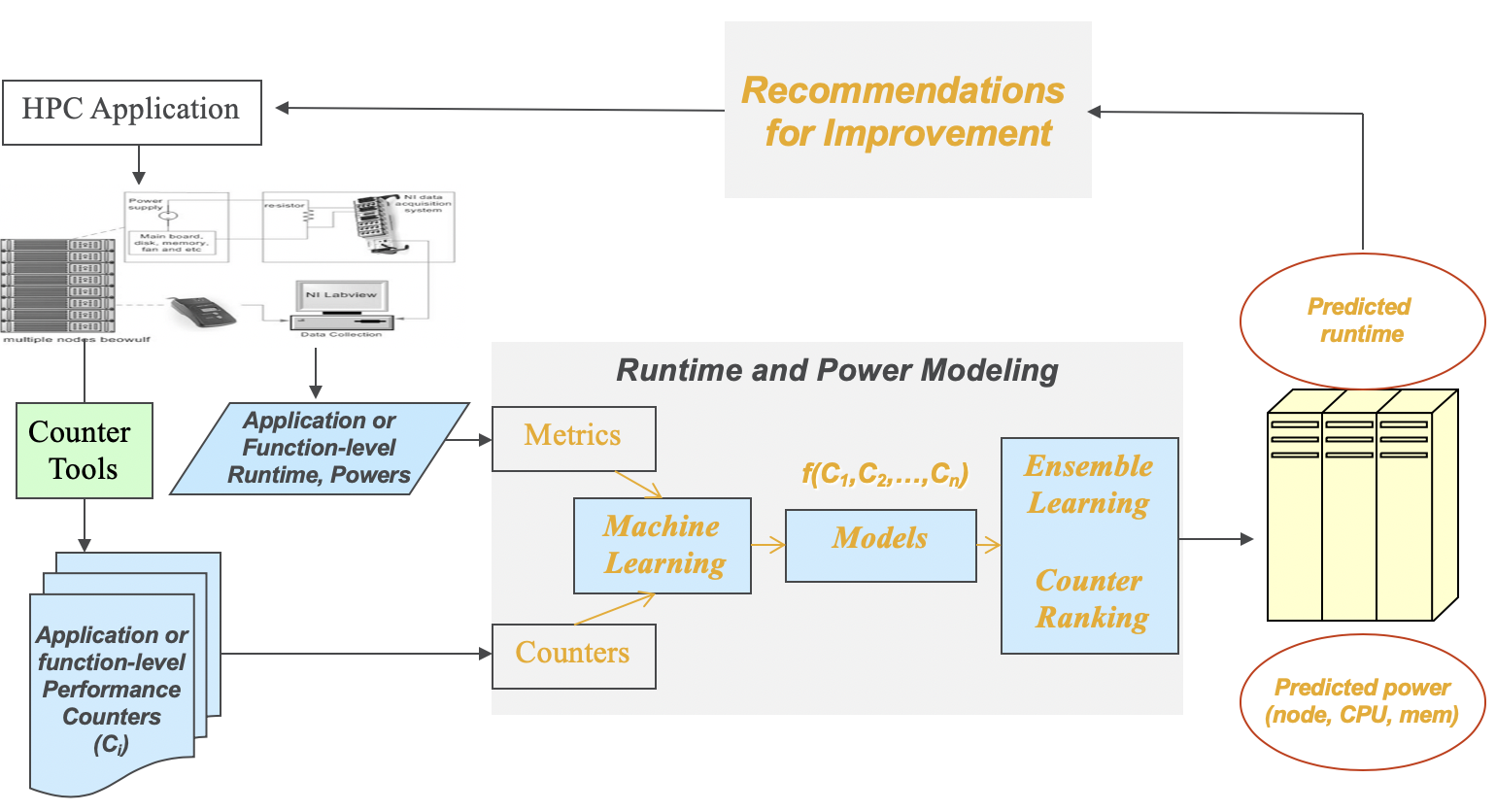}
 \caption{MuMMI Modeling and Improvement Framework}
\label{fig:1}
\end{figure}

Figure \ref{fig:1} shows the MuMMI modeling and improvement framework. This extends and enhances our previous MuMMI framework \cite{WT16} by leveraging ensemble learning and feature selection. For an HPC application executed on a power-aware system, we collect runtime, power (node, CPU, and memory), and performance counter data. During the application execution we capture available underived performance counters using any counter measurement tool (PAPI \cite{PAPI}, perf\_events \cite{PERF}, perfmon2 \cite{PMON}, HPM \cite{PO08}, etc.). All performance counters are normalized by using the total cycles of the execution to create performance event rates for each counter. We then use ML methods to build the models for the metrics and rank the counters based on their variable importance. However, each ML method builds the model and provides the variable importance in distinct ways. Hence, we use ensemble learning to to create the ensemble of these ML models, build a more accurate complex model, and provide the robust variable importance for ranking performance counters. Then we can use the counter ranking to identify the most important counters for potential application improvements and/or use the ensemble models to predict the execution time and power. In this work, we focus on four metrics: runtime, node power, CPU power, and memory power.

\section{Parallel Cancer Deep Learning CANDLE Benchmarks: P1B2 and NT3 }

In this section, we discuss two parallel caner deep learning CANDLE benchmarks \cite{21}: NT3 (weak scaling) and P1B2 (strong scling). The NT3 benchmark is a 1D convolutional network for classifying RNA-seq gene expression profiles into normal or tumor tissue categories. This network follows the classic architecture of convolutional models with multiple 1D convolutional layers interleaved with pooling layers followed by final dense layers. The model is trained on the balanced 700 matched normal-tumor gene expression profile pairs available from the NCI Genomic Data Commons and acts as a quality control check for synthetically generated gene expression profiles. The full dataset of expression features contains 60,483 float columns transformed from RNA-seq FPKM-UQ values \cite{21} that map to a column that contains the integer 0|1. The training data size for this benchmark is 597 MB, and the test data size is 150 MB. The number of epochs is one per node for the weak-scaling study. The batch size is 20 (default); and the total training samples are 1,120, with 60,483 features per sample. The optimizer is sgd (stochastic gradient descent).

The P1B2 benchmark is an MLP (multiLayer perceptron) network with regularization and five layers. Given patient somatic SNP data, it builds a deep learning network that can classify the cancer type based on sparse input data and evaluate the information content and predictive value in a molecular assay with auxiliary learning tasks. The training data size is 162 MB, and the test data size is 55 MB. The number of epochs is 384 for the strong-scaling study; the batch size is 60 (default); and the total training samples are 2,700 with 28,204 features per sample. The optimizer is rmsprop (root mean square propagation). 

In this work, we use the two benchmarks P1B2 and NT3  to collect the available 26 performance counters using PyPAPI \cite{PYPA} and performance and power data on the Cray XC40 Theta \cite{THET}. Each node of Theta has 64 cores. The dataset p1b2.csv for P1B2 has 144 configurations, which include 37 variables per configuration (such as application name and system name), 12 numbers of nodes (6 to 384), number of cores (384 to 24,576), learning rates, and 12 different batch sizes, number of epochs (384).
The dataset nt3.csv for NT3 has 105 configurations, which include the same 37 variables per configuration, 15 numbers of nodes (1 to 384), number of cores (64 to 24,576), learning rates, 7 different batch sizes, and number of epochs (one per node)
Both datasets include 26 performance counters and four metrics (runtime, node power, CPU power, and memory power). The number of cores used for P1B2 and NT3 is up to 24,576 (384 nodes). 

The 26 performance counters are TOT\_CYC (total cycles), TOT\_INS (total instructions completed), BR\_CN (conditional branch instructions), BR\_NTK (conditional branch instructions not taken), L1\_TCM (L1 total cache misses), L1\_LDM (L1 load misses), L1\_DCM (L1 data cache misses), L1\_ICA (L1 instruction cache accesses), L1\_ICH (L1 instruction cache hits), L1\_ICM (L1 instruction cache misses), L2\_TCM (L2 total cache misses), L2\_TCA (L2 total cache assesses), L2\_TCH (L2 total cache hits), L2\_LDM (L2 load misses), TLB\_DM (data translation lookaside buffer misses), BR\_MSP (conditional branch instructions mispredicted), RES\_STL (cycles stalled at any resource), SR\_INS (store instructions), LD\_INS (load instructions), BR\_TKN (conditional branch instructions taken), BR\_INS (branch instructions), L1\_DCA (L1 data cache accesses), LST\_INS (Load/store instructions completed), REF\_CYC (reference clock cycles), STL\_ICY (cycles with no instructions issued), and BR\_UCN (unconditional branch instructions).
Then TOT\_CYC is used to normalize all the performance counters. 
We used the datasets to analyze the pairwise correlations among 25 hardware performance counters and four target objects as follows.

\begin{figure}
\center
 \includegraphics[width=.45\textwidth]{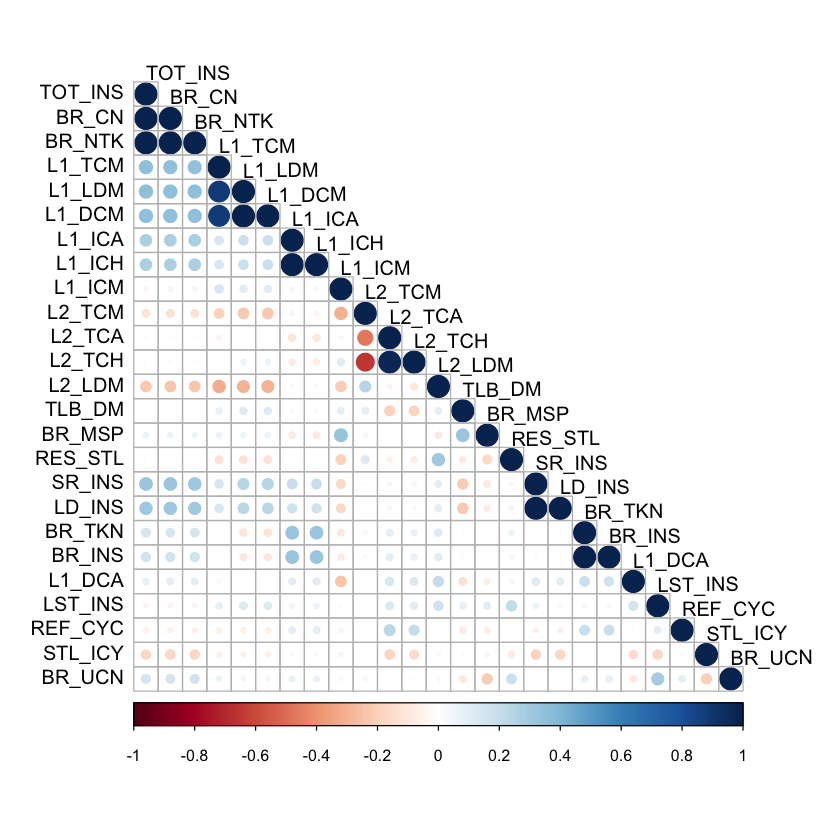}
 \caption{Counter correlation matrix for NT3}
\label{fig:2}
\end{figure}

\begin{figure}
\center
 \includegraphics[width=.45\textwidth]{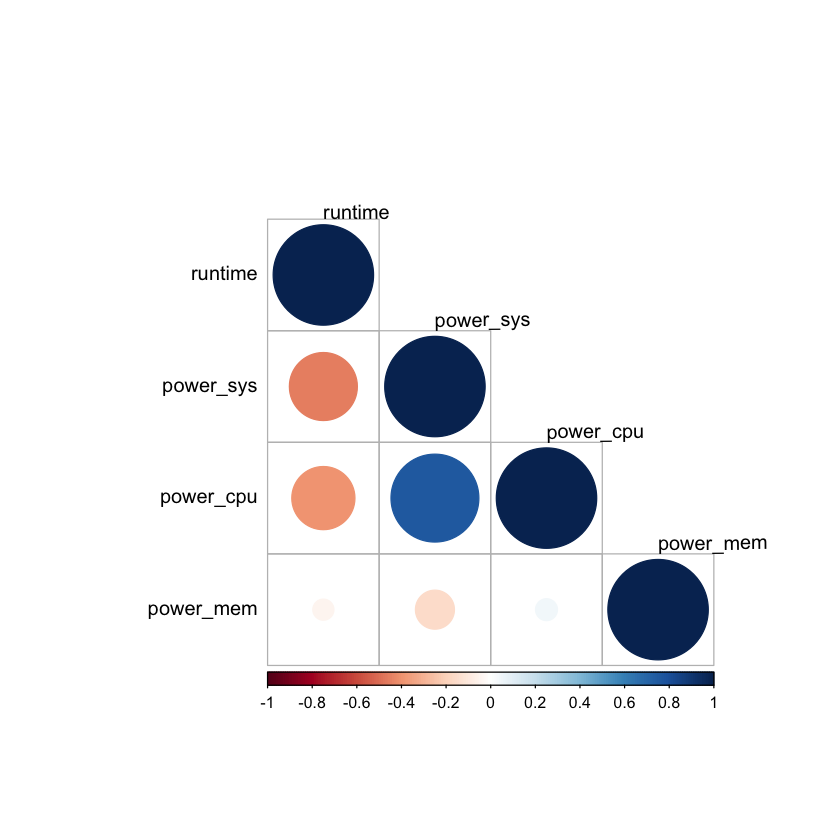}
 \caption{Object correlation matrix for NT3}
\label{fig:3}
\end{figure}

For NT3, we used weak scaling to collect the dataset nt3.csv with 105 configurations. For the same batch size, we expect that the runtime for different configurations is similar. Figure \ref{fig:2} shows the counter correlation matrix for 25 counters. This indicates that most counters are not correlated each other. Figure \ref{fig:3}  presents the object correlation matrix for the four metrics.
It indicates that runtime is inversely correlated with power. However, node power is highly correlated with CPU power because node power includes the CPU power, and CPU power is poorly correlated with memory power.

\begin{figure}
\center
 \includegraphics[width=.45\textwidth]{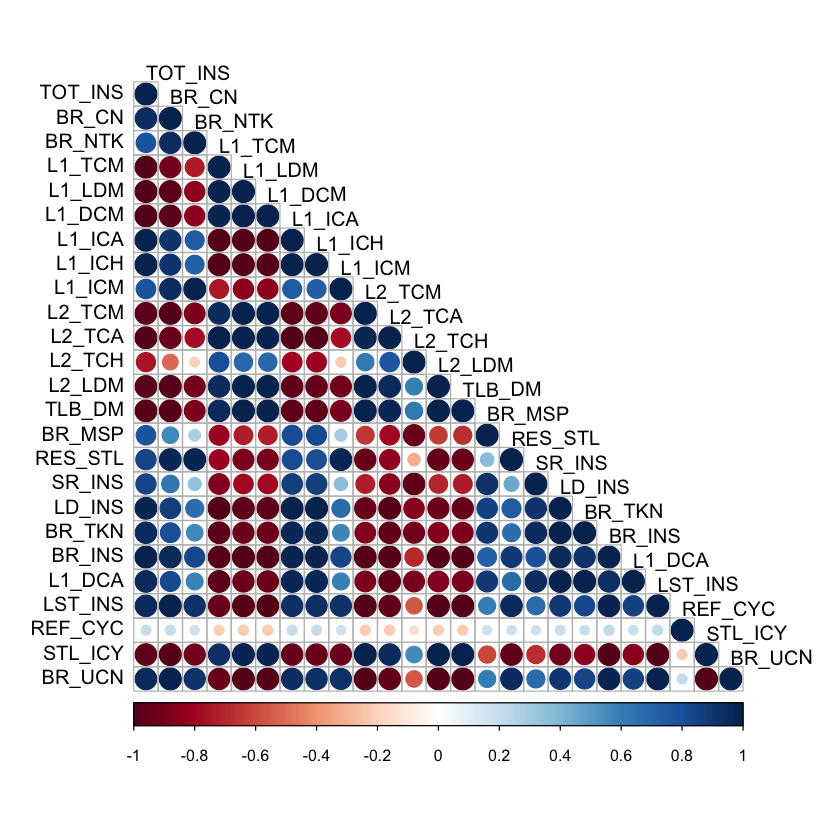}
 \caption{Counter correlation matrix for NT3}
\label{fig:4}
\end{figure}

\begin{figure}
\center
 \includegraphics[width=.45\textwidth]{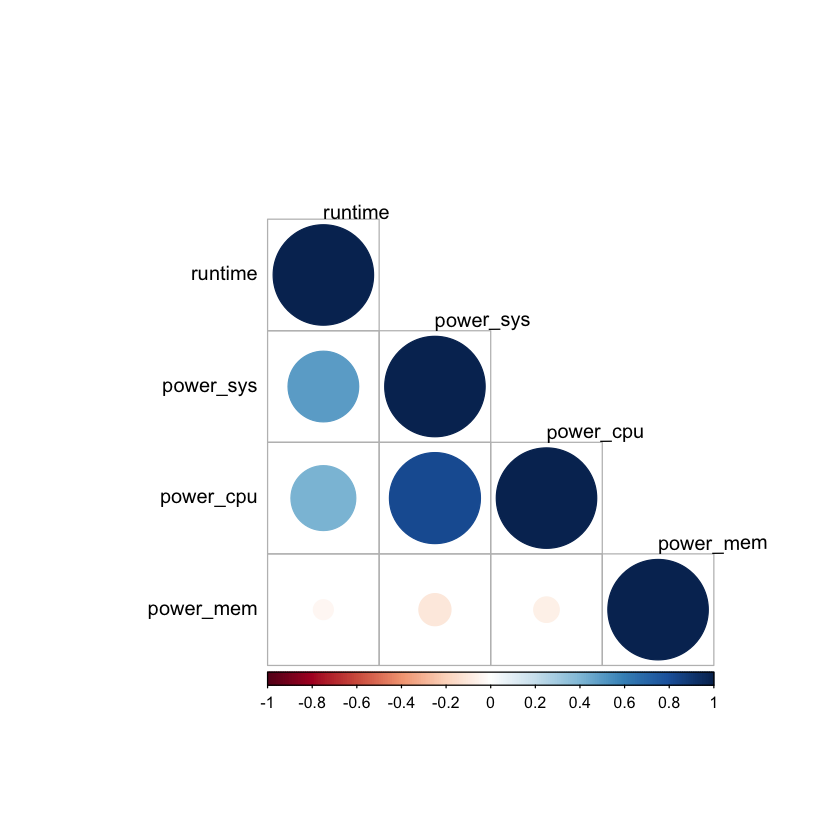}
 \caption{Object correlation matrix for NT3}
\label{fig:5}
\end{figure}

For P1B2, we used strong scaling to collect the dataset p1b2.csv with 144 configurations. For the same batch size, we expect that the runtime for different configurations is distinct and that it decreases with increasing numbers of nodes because the number of epochs per node is decreased. Figure \ref{fig:4}  shows the counter correlation matrix for 25 counters. This indicates that most counters have some correlation each other except REF\_CYC. Figure \ref{fig:5}  presents the metrics correlation matrix for four metrics.
It indicates that runtime is correlated with power.
As with NT3, however, node power is highly correlated with CPU power because node power includes the CPU power, and CPU power is poorly correlated with memory power.

\section{Performance and Power Modeling Using Ensemble Learning}

To make sure we use the same training and test sets, we use the set.seed(3456) in our R codes for all experiments so that creating the random objects can be reproduced. We define the relationship between the object metric (runtime, node power, CPU power, or memory power) and variables (25 performance counters) as follows:
metric  $\sim $ TOT\_INS + BR\_CN + BR\_NTK + L1\_TCM + L1\_LDM + L1\_DCM + L1\_ICA + L1\_ICH + L1\_ICM + L2\_TCM + L2\_TCA + L2\_TCH + L2\_LDM + TLB\_DM + BR\_MSP + RES\_STL + SR\_INS + LD\_INS + BR\_TKN + BR\_INS + L1\_DCA  + LST\_INS + REF\_CYC + STL\_ICY + BR\_UCN.

When using models to predict a numeric outcome, some measure of accuracy is typically used to evaluate the effectiveness of the model. For instance, the most common method for characterizing a model's predictive capabilities is to use the root mean square error (RMSE). RMSE is a function of the model residuals, which are computed as the observed values minus the predicted value. It is interpreted  either as how far, on average, the residuals are from zero or as the average distance between the observed values and the predicted values. Another common metric is $R^2$,  the coefficient of determination, which can be interpreted as the proportion of the information in the data that is explained by the model. It is the squared correlation coefficient between the observed and predicted values in the simplest version. It is a measure of correlation, not accuracy. Therefore, in this paper we use the metric RMSE to evaluate the effectiveness of the model using different ML methods and ensemble learning.

We choose 15 machine learning methods \cite{1} from the R caret package:  tree-/rule-based group -- random forest (rf), cubist (cubist), eXtreme Gradient Boosting (xgbTree), conditional inference tree (ctree), and treebag; nonlinear group -- k-nearest neighbors (knn), support vector machines with linear kernel (svmLinear), Gaussian process with radial basis function (gaussprRadial), multivariate adaptive regression spline (earth), and Bayesian generalized linear model (bayesglm); linear group -- Lasso and elastic-net regularized generalized linear models (glmnet), partial least squares (pls), ridge fegression (ridge), elastic vet regression (enet), and principal component regression (pcr) because their RMSE is very close to each other in the same group. Then we use caretEnsemble to create the ensemble of the 15 ML methods and build the model and rank the performance counters.

\subsection{Performance and Power Modeling}

In this section, we use the 15 machine learning methods and caretEnsemble \cite{MZ16} to model performance (runtime) and node power of NT3 and P1B2 using the same training and test sets based on the 80/20\% rule. 

Table \ref{tab:1} shows the ensemble model for P1B2 that resulted from the ensemble of the following models: cubist, treebag, xgbTree, ctree, rf, glmnet, pls, ridge, pcr, enet, knn, svmLinear, earth, gaussprRadial, and bayesglm. The resulting RMSE for the ensemble model is 200.14 in performance and 3.70 in node power. They are the most accurate among these models.

\begin{table}
\center
\caption{RMSE of different ML models for P1B2}
\begin{tabular}{c}
  \includegraphics[width=.45\textwidth]{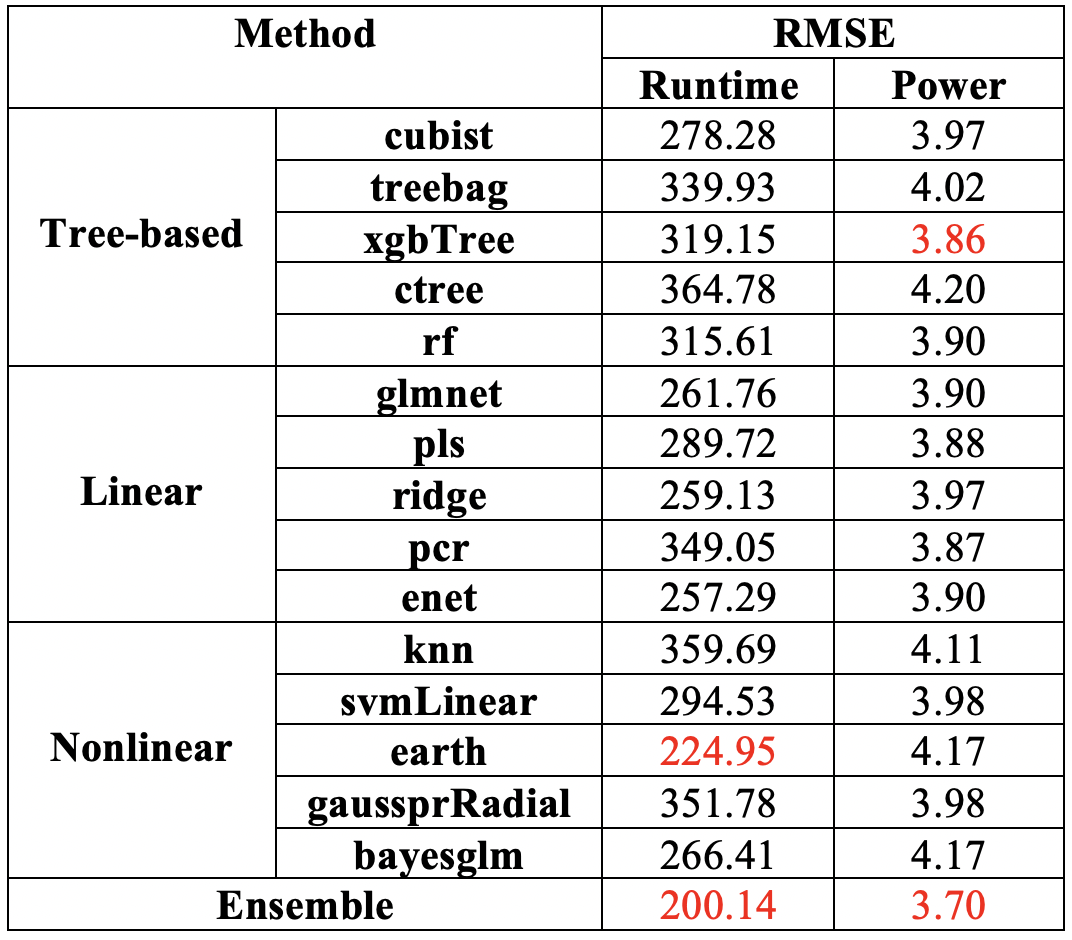}
  \end{tabular}
\label{tab:1}       
\end{table}  

Similarly, Table \ref{tab:2} shows the ensemble model for NT3 thst resulted from the ensemble of the following models: cubist, treebag, xgbTree, ctree, rf, glmnet, pls, ridge, pcr, enet, knn, svmLinear, earth, gaussprRadial, and bayesglm. The resulting RMSE for the ensemble model is 154 in performance and 4.24 in node power. The ensemble model for runtime is the most accurate among these models; however, the power model using rf results in the smallest RMSE, 4.22. This is an exception. Overall, ensemble learning results in the most accurate performance and power models for P1B2 and NT3 in most cases.

\begin{table}
\center
\caption{RMSE of different ML models for NT3}
\begin{tabular}{c}
  \includegraphics[width=.45\textwidth]{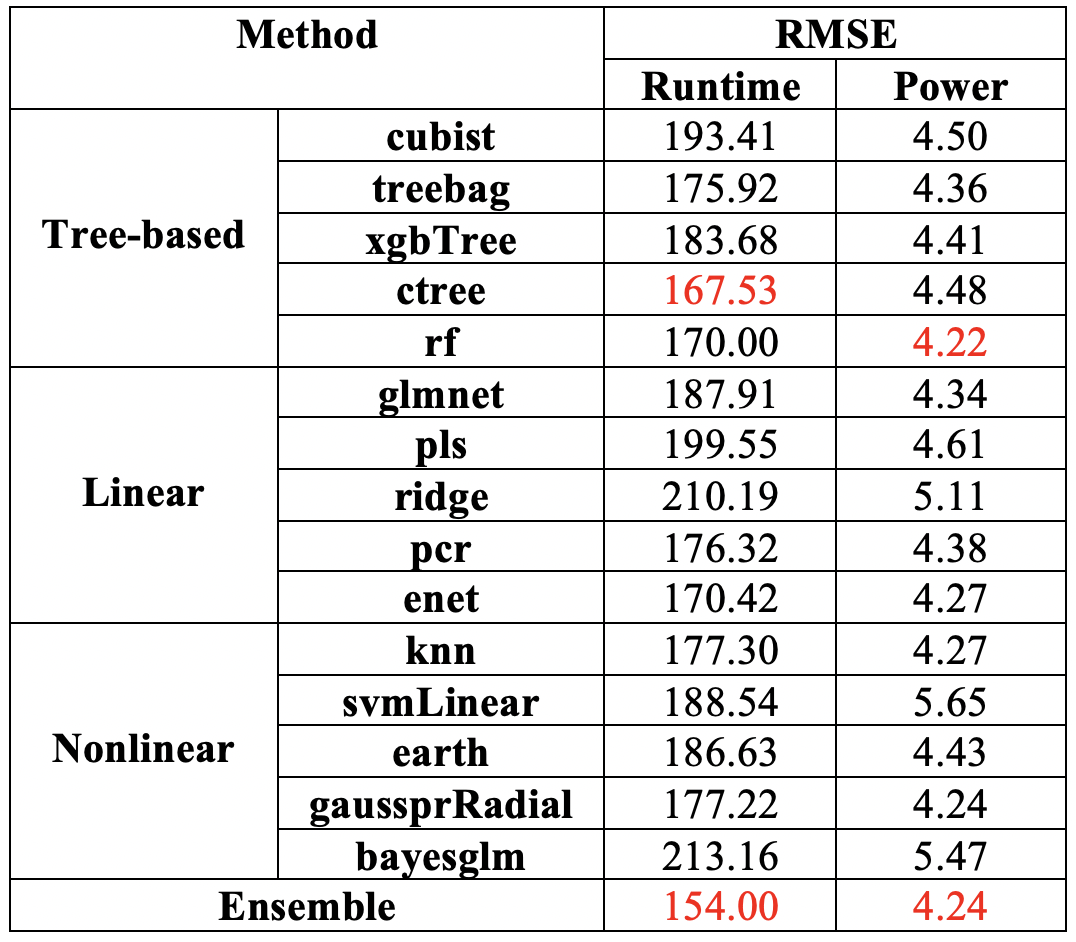}
  \end{tabular}
\label{tab:2}       
\end{table}  

\subsection{Performance Counter Ranking}

In this section, we use P1B2  to explore performance counter ranking using model-based feature selection for 15 ML methods and ensemble learning and using unsupervised feature selection. We compare them to identify the most important performance counters that impact the performance and node power. 

\subsubsection{Model-Based (Supervised) Feature Selection}

In this section, we use 15 machine learning methods and caretEnsemble to explore performance counter ranking. varImp() is used to extract the variable importance from each member of the ensemble, as well as the final ensemble model. Then we sum the variable importance values for counters to calculate the percentage for each counter to rank them. 

\begin{table}
\center
\caption{Performance counter ranking for performance models}
\begin{tabular}{c}
  \includegraphics[width=.48\textwidth]{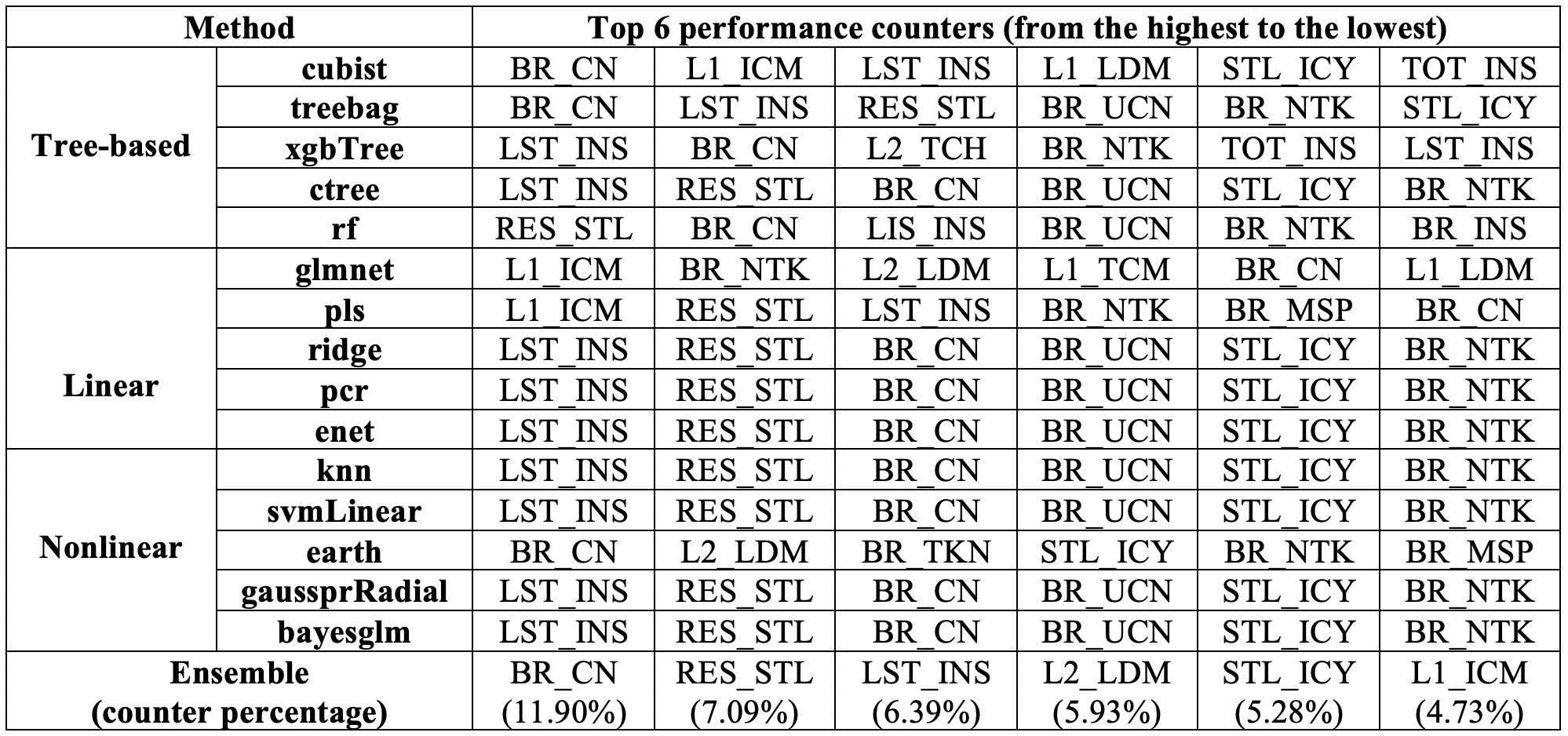}
  \end{tabular}
\label{tab:3}       
\end{table}

\begin{figure}
\center
 \includegraphics[width=.45\textwidth]{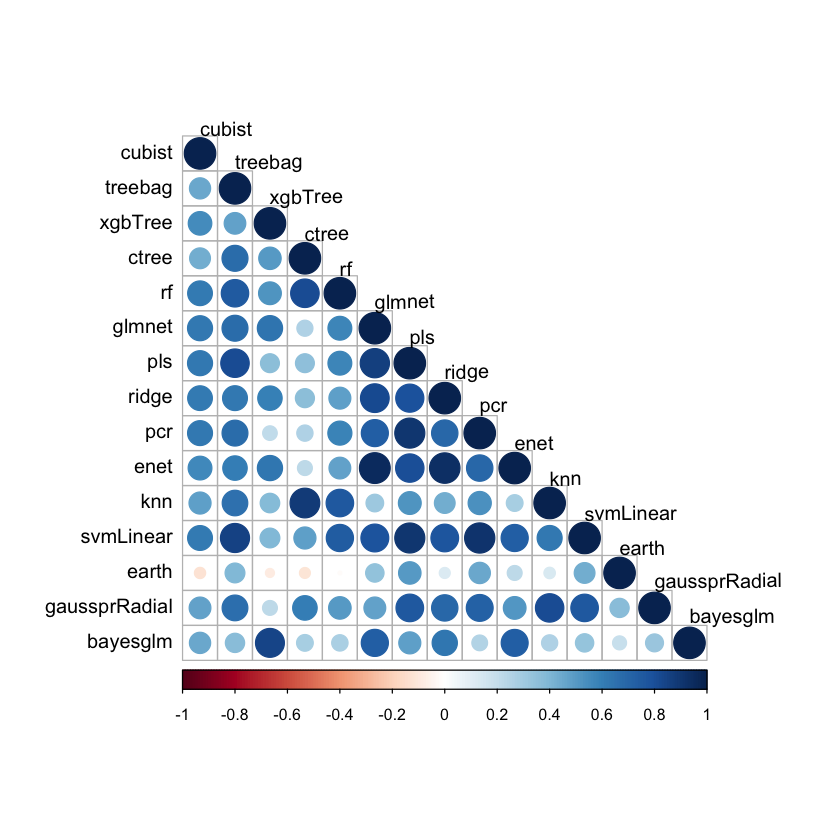}
 \caption{ML performance model correlation matrix for P1B2}
\label{fig:6}
\end{figure}

Table \ref{tab:3} presents the top 6 counters in each performance model using the different ML methods. 
A total of 15 different counters are listed in the table. 
We observe that 7 of the 15 ML methods listed (ridge, pcr, enet, knn, svmLinear, gaussprRadia, and bayesglm) have the same performance counter ranking order. However, they are not highly correlated each other, as shown in Figure \ref{fig:6}. The others have different performance counter rankings. We note that ensemble learning provides the overall counter ranking, which is different from all 15 ML methods and includes the top counter from each ML model. The counter percentage is the counter importance values divided by the summation of all counter importance values. We note that BR\_CN, RES\_STL, and LST\_INS are the top 3 counters in performance modeling provided by ensemble learning. 

\begin{figure}
\center
 \includegraphics[width=.45\textwidth]{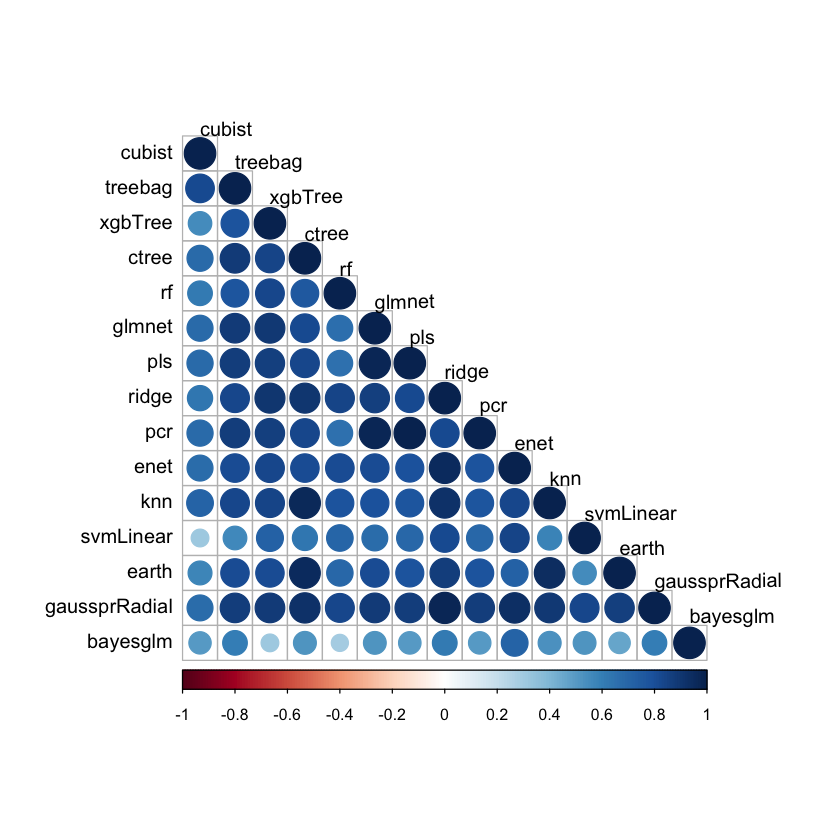}
 \caption{ML power model correlation matrix for P1B2}
\label{fig:7}
\end{figure}

Table \ref{fig:4} presents the top 6 counters in each node power model using  different ML methods, which are different from that in Table \ref{tab:3}. A total of 21 different counters are listed in the table. We observe that the same 7 ML methods (ridge, pcr, enet, knn, svmLinear, gaussprRadia, and bayesglm) have the same performance counter ranking order and that ridge and gaussprRadia have a high correlation with the others, as shown in Figure \ref{fig:7}. We note that L1\_ICH, LST\_INS, and L2\_LDM are the top 3 counters in power modeling provided by ensemble learning. 

\begin{table}
\center
\caption{Performance counter ranking for power models}
\begin{tabular}{c}
  \includegraphics[width=.48\textwidth]{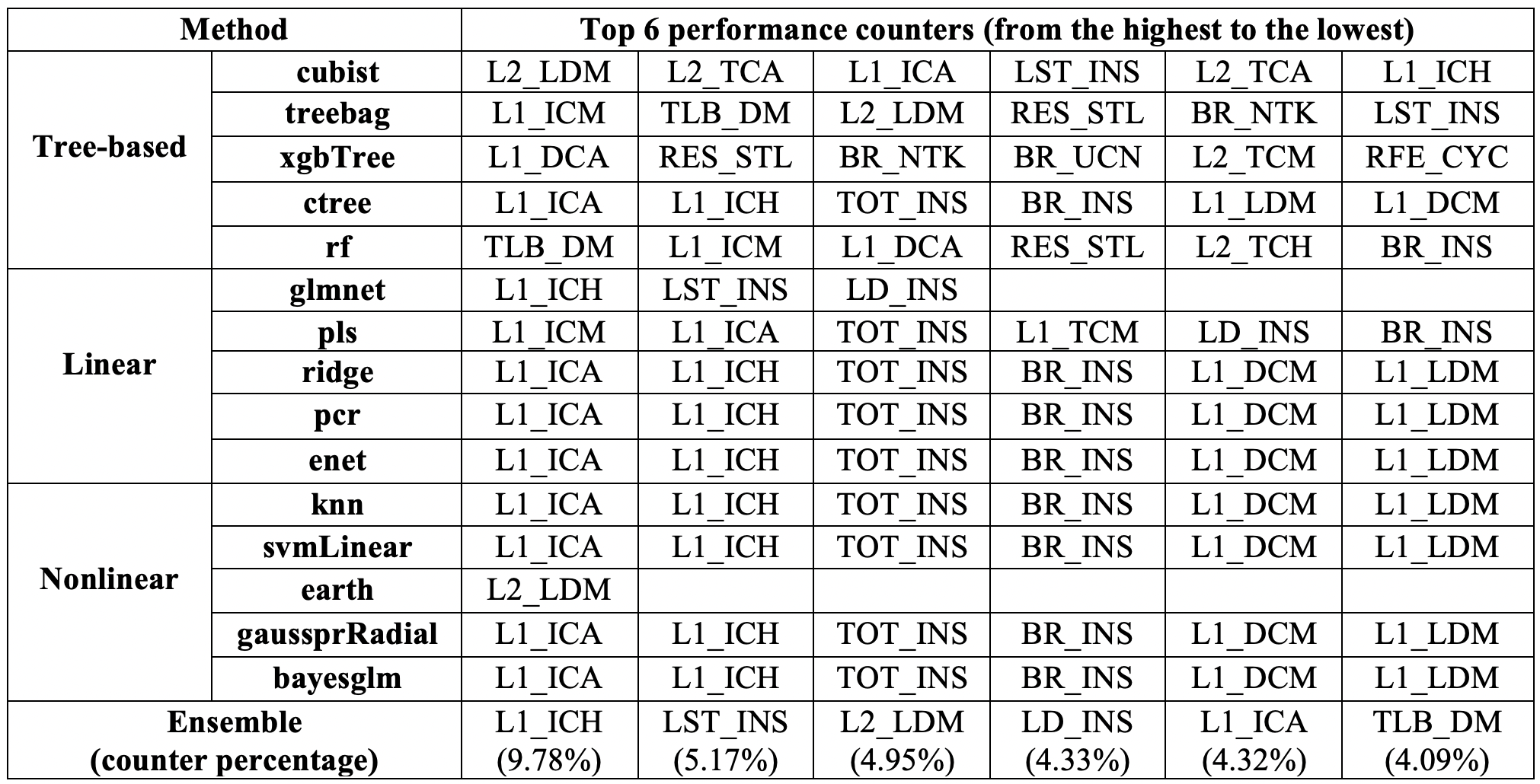}
  \end{tabular}
\label{tab:4}       
\end{table}

\subsubsection{Unsupervised Feature Selection}

Next, we use unsupervised feature selection and compare with the model-based feature selection. For unsupervised feature selection, we use two types of methods \cite{KJ13}: (1) wrapper methods -- recursive feature elimination (RFE), genetic algorithm (GA), and simulated annealing (SA); and (2) filter methods -- stepwise and selection by filter (SBF). Then we compare them with the results from the ensemble learning. We still use P1B2 as an example.

For the RFE method, it supports random forest (rfFuncs), linear regression (lmFuncs), and bagged tree (treebagFuncs). For the GA method, it supports random forest (rfGA) and bagged tree (treebagGA). For the SA method, it supports random forests (rfSA) and bagged tree (treebagSA). For the SBF method, it supports random forests (rfSBF), linear regression (lmSBF), and bagged tree (treebagSBF). Stepwise forward and backward selection use glm. 

Table \ref{tab:5} shows the performance counter ranking for performance models using unsupervised feature selection. It confirms that the top two counters are BR\_CN and RES\_STL provided by ensemble learning in Table \ref{tab:3}. The other counters also occur in Table \ref{tab:5}.

\begin{table}
\center
\caption{Performance counter ranking for performance models}
\begin{tabular}{c}
  \includegraphics[width=.45\textwidth]{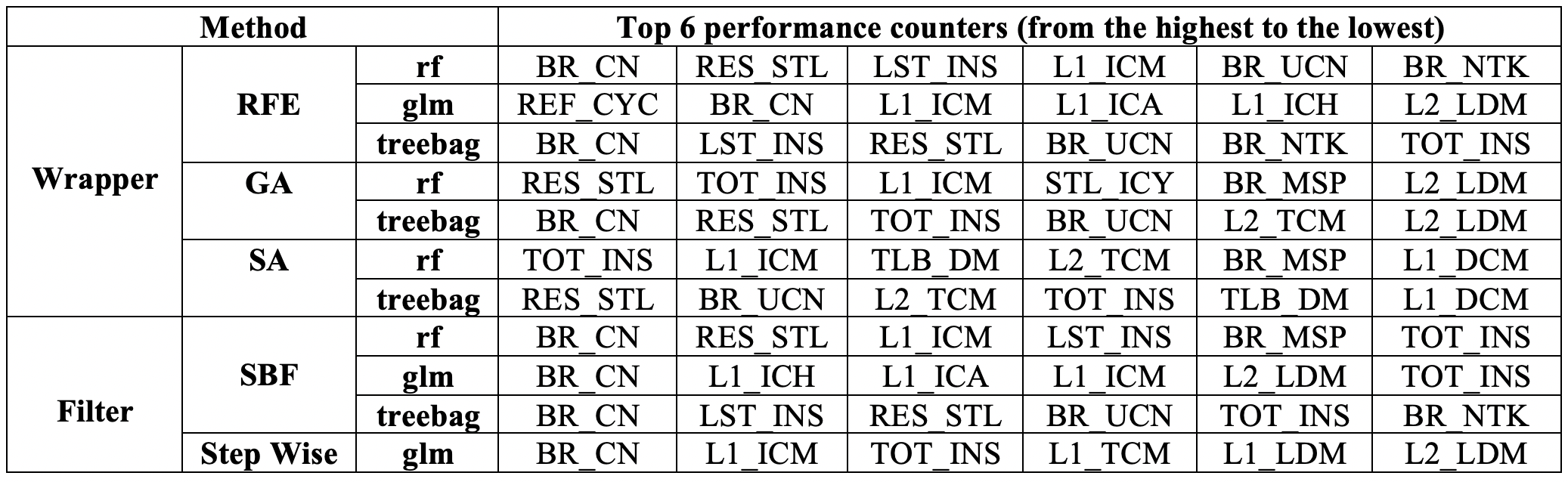}
  \end{tabular}
\label{tab:5}       
\end{table}

\begin{table}
\center
\caption{Performance counter ranking for power models}
\begin{tabular}{c}
  \includegraphics[width=.45\textwidth]{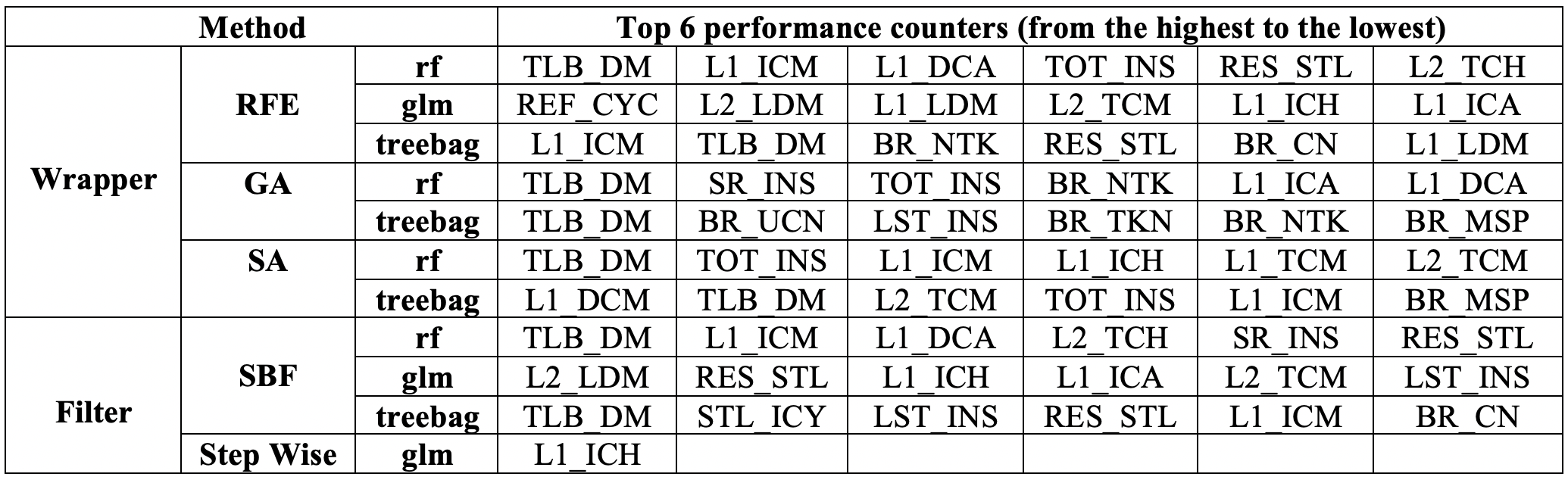}
  \end{tabular}
\label{tab:6}       
\end{table}

Table \ref{tab:6} shows the performance counter ranking for node power models using unsupervised feature selection. It shows that TLB\_DM is the top counter in  the top 6 counters in Table \ref{tab:4}.  Two of the top three counters, L1\_ICH, LST\_INS, and L2\_LDM, provided by ensemble learning in Table \ref{tab:4} are the top counters in Table \ref{tab:6}. Overall, compared with each  ML method, we find that ensemble learning provides more robust performance count ranking. 

\subsection{Discussion}

In summary, for P1B2, we used unsupervised feature selection methods to confirm that ensemble learning provides more robust performance count ranking than each individual ML method does. 
When we consider runtime and node power for application improvement for P1B2, we need to focus on the counters BR\_CN, L1\_ICH, and TLB\_DM. Similarly, when we apply the same methodology to NT3, we find that the counters L1\_ICM, L2\_TCH, and TLB\_DM are the focal counters for application improvement.

\section{Multiple-Objects Ensemble Learning}

In the preceding sections, we explored single-object (univariate) ensemble learning \cite{MZ16} to combine different ML methods and result in more accurate models and provide more robust performance counter ranking. These methods  target only a single metric, either runtime or node power. In this section, we discuss multiple-objects (multivariate) ensemble learning using a multivariate tree boosting method mvtboost \cite{MP16} to model performance and power and to rank the performance counters based on multiple objects. 

Boosted decision tree ensembles such as gradient boosting machine (gbm) \cite{FJ01} are powerful ensemble learning algorithms, allowing dependent variables to be nonlinear functions of predictors. mvtboost (multivariate tree boosting) extends gbm to multivariate, continuous object variables by fitting a separate univariate model of a common set of predictors to each object variable. This accounts for covariance in the object variables as in seemingly unrelated regression. This joint analysis of several object variables can be informative when we consider the four metrics runtime, node power, CPU power, and memory power for application improvement. In this section, we use a mvtboost method with 1,000 trees, a learning rate of 0.01, and a tree depth of 3 to model the performance and power of P1B2 and NT3 and analyze the impacts of different performance counters. We then compare the performance counter ranking with what we found in the preceding section. 

Table \ref{tab:7} shows the performance counter ranking for P1B2 using multiple-objects ensemble learning. When we consider runtime, node power, CPU power, and memory power for the application improvement, we need to focus on the top counters BR\_CN, TLB\_DM, and L1\_ICM,  similar to the counters BR\_CN, L1\_ICH, and TLB\_DM found in the preceding section. This further confirms that we need to focus on BR\_CN, L1 cache, and TLB\_DM to improve the application performance and power.

\begin{table}
\center
\caption{Performance counter ranking for P1B2 using ensemble learning}
\begin{tabular}{c}
  \includegraphics[width=.45\textwidth]{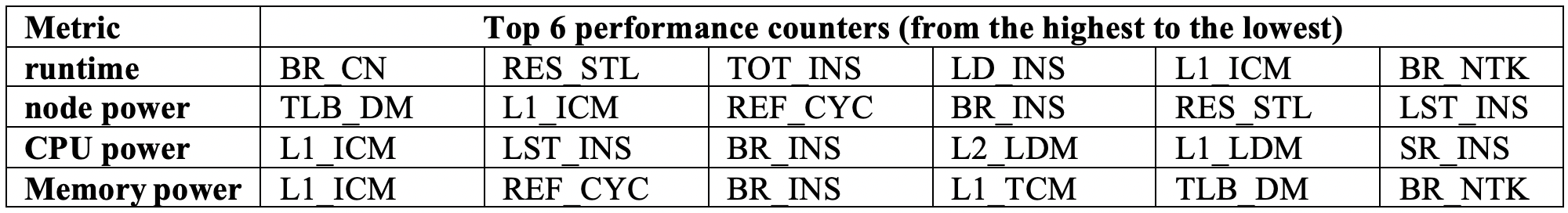}
  \end{tabular}
\label{tab:7}       
\end{table}

\begin{table}
\center
\caption{Performance counter ranking for NT3 using ensemble learning}
\begin{tabular}{c}
  \includegraphics[width=.45\textwidth]{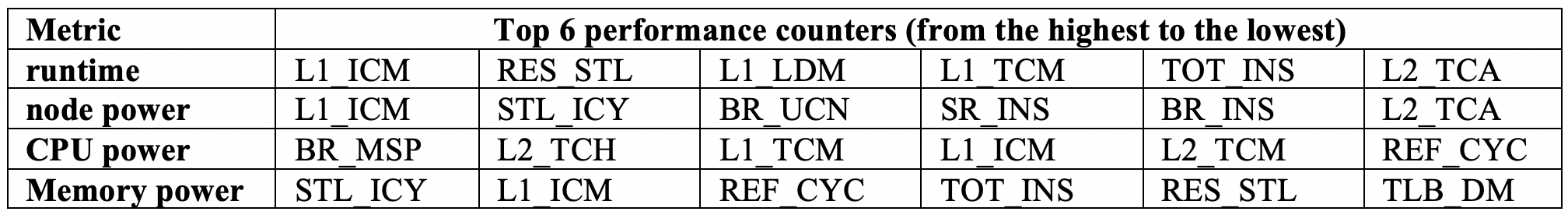}
  \end{tabular}
\label{tab:8}       
\end{table}

Table \ref{tab:8} shows the performance counter ranking for NT3 using multiple-objects ensemble learning. When we consider runtime, node power, CPU power, and memory power for the application improvement, we need to focus on the top counters L1\_ICM, BR\_MSP, and STL\_ICY. In the preceding section, when we consider runtime and node power, we need to focus on L1\_ICM, L2\_TCH, and TLB\_DM. Overall, we need to focus on L1 cache, BR\_MSP, STL\_ICY, and TLB\_DM to improve the application performance and power.

\section{Performance and Energy improvement }

The CANDLE benchmarks P1B2 and NT3 are implemented in Python by using  Keras, TensorFlow, and Horovod. 
These Python codes, like other scripting languages, do not have compiler optimization support and instead rely on the library, resource, and environment settings for better performance. We can utilize what we learned from ensemble learning in preceding sections to identify better resource and environment settings for performance improvement. In this section, based on the insights from important performance counters we identified, we  improve the performance and energy of these benchmarks by fine-tuning the applications, underlying library, software, and/or system. 

For P1B2 and NT3 executed on Theta, the dominant phase is model training \cite{WT19}, which is the function call from TensorFlow. From our analysis in the preceding sections, the most important counters indicate that TensorFlow is the optimization target with  huge page sizes. The Cray XC40 Theta \cite{THET} supports the huge page sizes of 2 MB, 4 MB, 8 MB, 16 MB, 32 MB, 64 MB, 12 8MB, 256 MB, 512 MB, 1 GB, and 2 GB. The baseline environment used in the work includes Python3.7.5, TensorFlow 1.15, Keras 2.3, and Horovod 0.18 built from the packages by using pip. To optimize TensorFlow with the huge page sizes, we have to rebuild from the source codes for TensorFlow 1.15, Keras 2.3, and Horovod 0.19 based on Intel Python 3.6.8 and MKL-DNN with  page sizes of 2 MB, 8 MB, 32 MB, 128 MB, and 1 GB. Then we use the benchmarks under these different environments to evaluate performance and energy. After we identify which page size results in the best performance, we further improve the application performance and energy.

\subsection{Impacts of Different Huge Page Sizes}

In this section, we use P1B2 and NT3 to evaluate the impacts of different huge page sizes of 2 MB, 8 MB, 32 MB, 128 MB, and 1 GB. For P1B2, we use a strong-scaling study (384 epochs in total) with the batch size of 60 to analyze the improvement. For NT3, we use a weak-scaling study (one epoch per node) with a batch size of 20 to analyze the improvement.
 
Table \ref{tab:9} shows performance and energy improvement for NT3 under different huge page sizes, where "2MB" stands for  the huge page size of 2 MB. 
We rebuilt the deep learning environment from the source codes for TensorFlow 1.15, Keras 2.3, and Horovod 0.19 based on Intel Python 3.6.8 and MKL-DNN. Table \ref{tab:10} shows performance and energy improvement for P1B2 under different huge page sizes. For both benchmarks, as we showed in \cite{WT19}, the modeling training and data loading are the two dominant phases. Compared with the baseline, we find that building TensorFlow from source codes based on the latest MKL-DNN improves the model training performance significantly, and the huge page size of 8 MB outperforms the others in most cases. However, the data-loading time under the baseline with Python 3.7.5 (around 115 s for NT3, around 105 s for P1B2) is much smaller than that (around 160 s for NT3, around 125 s for P1B2) under the different huge page sizes with Python 3.6.8. We used the improved data loading methods from \cite{WT19} for both cases. Basically, the data loading calls the Pandas function read\_csv() to load the data. For all cases, the installed Pandas versions are the same (version 0.25.2). The difference is the Python version. In the next section, we  investigate this issue to further improve the performance and energy.

\begin{table}
\center
\caption{Performance and energy improvement for NT3}
\begin{tabular}{c}
  \includegraphics[width=.45\textwidth]{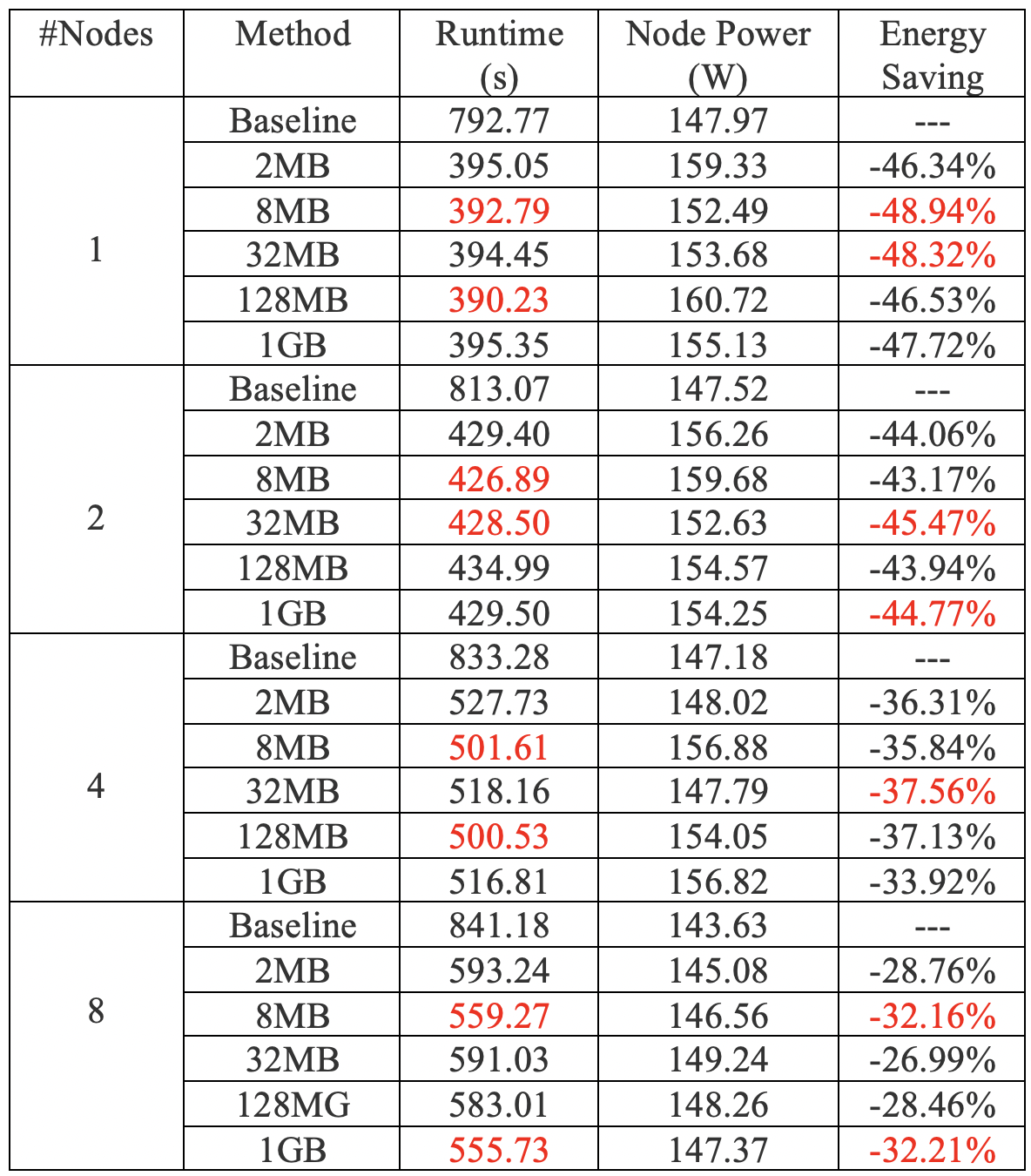}
  \end{tabular}
\label{tab:9}       
\end{table}  

\begin{table}
\center
\caption{Performance and energy improvement for P1B2}
\begin{tabular}{c}
  \includegraphics[width=.45\textwidth]{imp-nt3.png}
  \end{tabular}
\label{tab:10}       
\end{table}  

XLA (Accelerated Linear Algebra) \cite{XLA} is a domain-specific compiler for linear algebra that can accelerate TensorFlow models with potentially no source code changes. It compiles subgraphs to reduce the execution time of short-lived operations to eliminate overhead from the TensorFlow runtime, fuses pipelined operations to reduce memory overhead, and specializes to known tensor shapes to allow for more aggressive constant propagation. It analyzes and schedules memory usage, in principle eliminating many intermediate storage buffers. As described in \cite{XLA}, XLA may improve the execution speed and memory usage on GPUs. However, it does not support CPUs well, as we experienced on Theta. Table \ref{tab:11} shows the XLA impacts using 8 nodes on Theta. For NT3 and P1B2, we enable XLA to rebuild TensorFlow to run them with/without enabling autoclustering by setting the TF\_XLA\_FLAGS  environment variable as ``--tf\_xla\_auto\_jit=2  --tf\_xla\_cpu\_global\_jit'' under the huge page size of 8 MB. Enabling XLA caused runtime to increase significantly; however, it results in the average node power decrease because of the XLA overhead. This is the case for all other experiments as well.

\begin{table}
\center
\caption{XLA impacts using 8 nodes under the huge page size of 8 MB on Theta}
\begin{tabular}{c}
  \includegraphics[width=.45\textwidth]{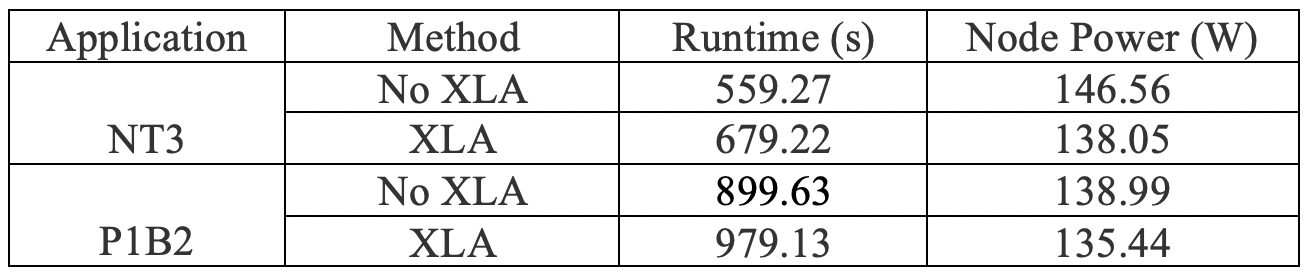}
  \end{tabular}
\label{tab:11}       
\end{table}

\subsection{Further Improvement}

For further improvement under the huge page size of 8 MB, we also rebuilt Python3.7.5, TensorFlow 1.15, Keras 2.3, and Horovod 0.19 from the source codes based on the latest Intel MKL-DNN. We then evaluated the performance and energy and compared them with "Baseline" and "2MB'', as described in the preceding section. 

Table \ref{tab:12} shows the performance and energy improvement for NT3 on up to 384 nodes (24,576 cores) under three different deep learning environments. We note that for the weak-scaling case study of NT3, the workload per node is the same (1 epoch per node), and we focus on improving the TensorFlow training performance. With an increase in the number of nodes, the overall energy saving percentage decreases because of the increase of Horovod communication overhead. This overhead also results in a decrease in the average node power. Overall, we use the further improvement environment to achieve up to 55.81\% performance improvement and up to 52.60\% energy saving on up to 24,576 cores. Compared with the "2MB'' and the baseline, we further improve not only the model training time but also the data-loading time. Thus, we achieve much better energy saving.

\begin{table}
\center
\caption{Performance and energy improvement for NT3 on up to 24,576 cores}
\begin{tabular}{c}
  \includegraphics[width=.45\textwidth]{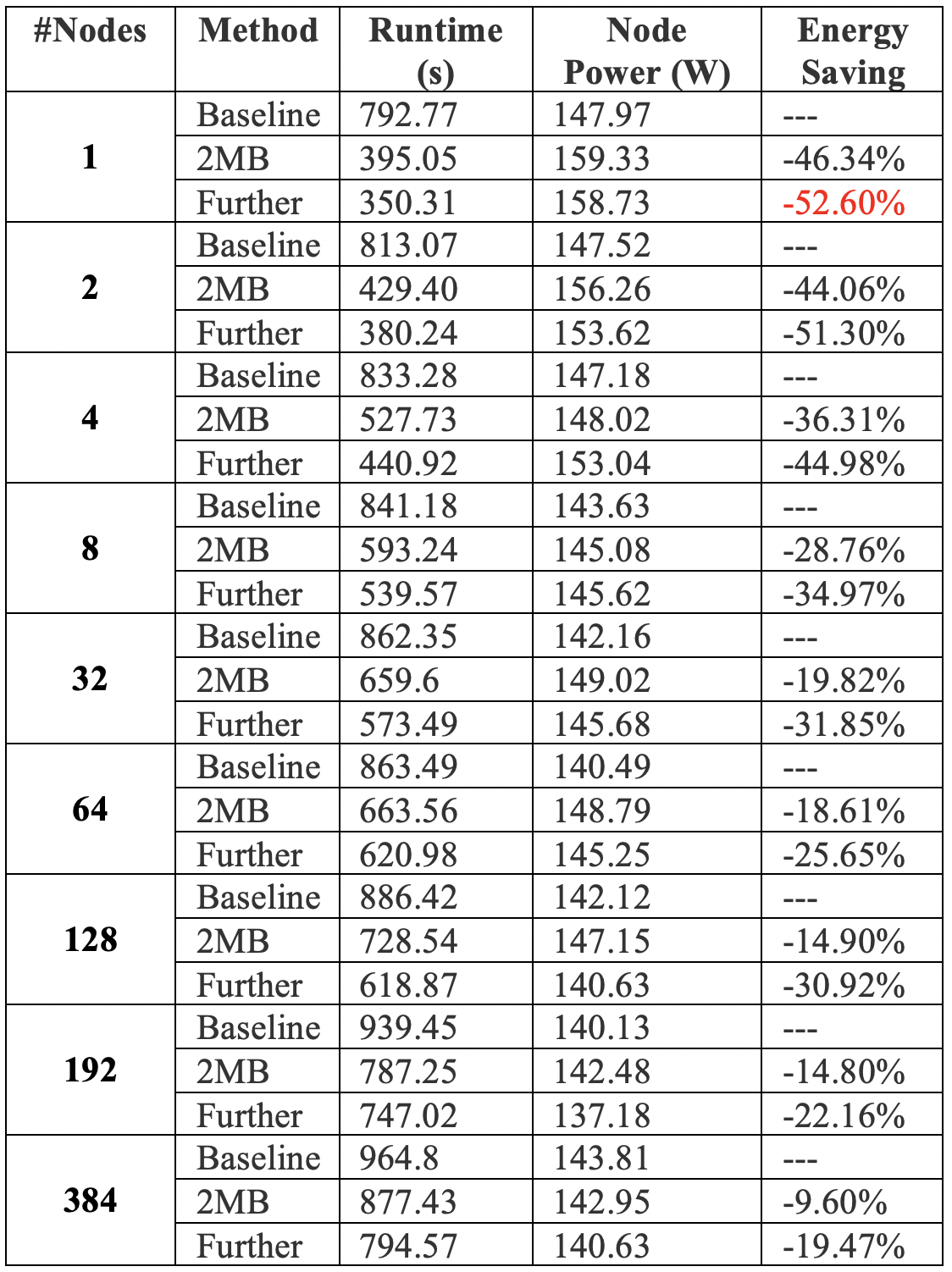}
  \end{tabular}
\label{tab:12}       
\end{table}

\begin{table}
\center
\caption{Performance and energy improvement for P1B2}
\begin{tabular}{c}
  \includegraphics[width=.45\textwidth]{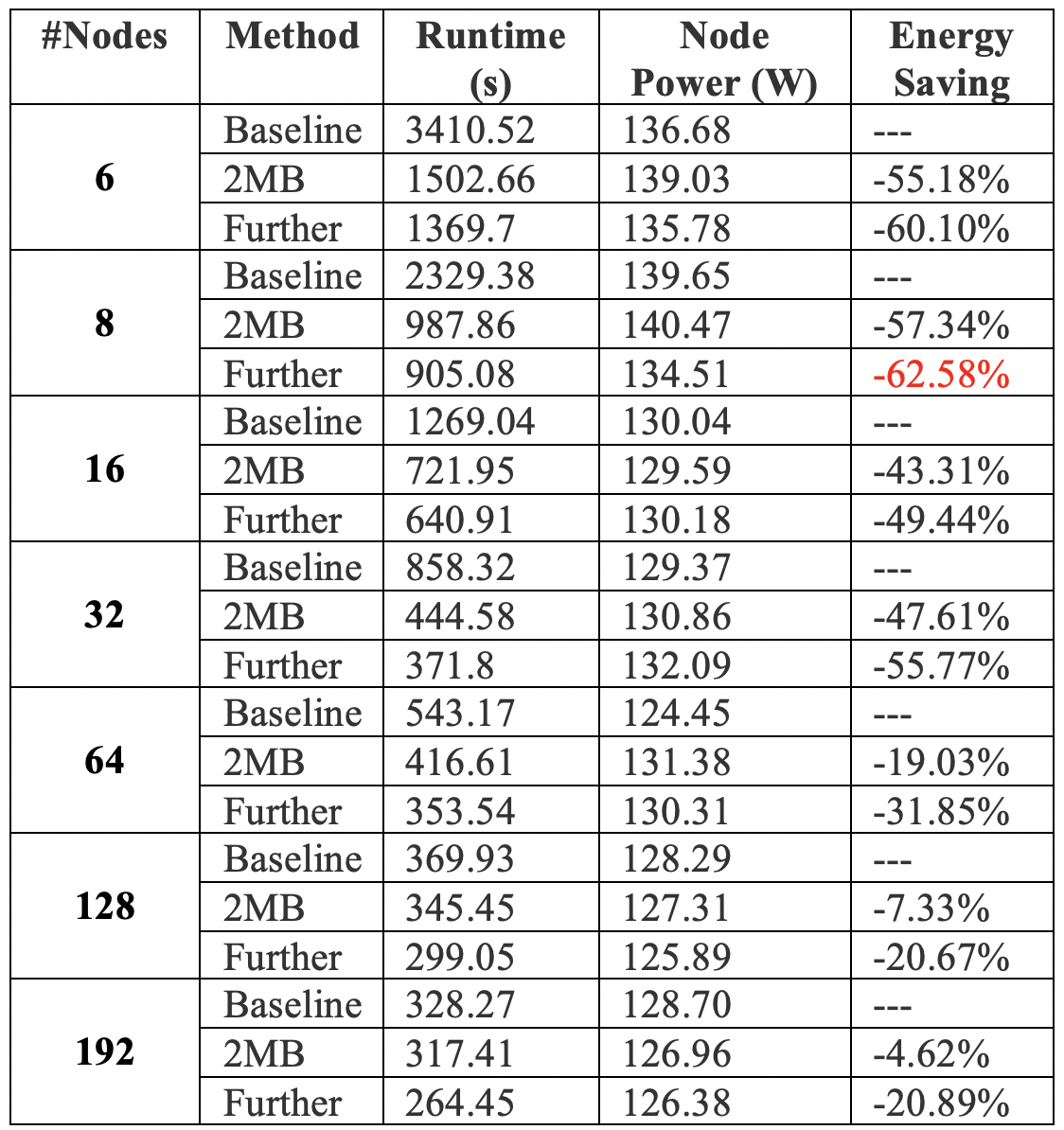}
  \end{tabular}
\label{tab:13}       
\end{table}

Table \ref{tab:13} shows the performance and energy improvement for P1B2 on up to 192 nodes (12,288 cores) under three different deep learning environments. We note that for the strong-scaling case study of P1B2, the workload per node decreases with increasing numbers of nodes. We focus on improving the TensorFlow training performance. With the increase in the number of nodes, the overall energy saving percentage decreases because of the decrease in the workload per node and the increase in the Horovod communication overhead. This also results in a decrease in the average node power. Overall, we use the further improvement environment to achieve up to 61.15\% performance improvement and up to 62.58\% energy saving on up to 12,288 cores.

\section{Conclusions}

We used the datasets collected for the two benchmarks NT3 (weak scaling) and P1B2 (strong scaling) to build performance and power models based on hardware performance counters using single-object and multiple-object ensemble learning. We utilized ensemble learning to combine linear, nonlinear, and tree-/rule-based ML methods to cope with the bias-variance tradeoff and result in more accurate models; and we ranked the performance counters to identify the most important counters for improvement hints. Based on the insights from these models, we improved the performance and energy of P1B2 and NT3 by optimizing the deep learning environments  TensorFlow, Keras, Horovod, and Python under the huge page size of 8 MB on Theta. Experimental results show that ensemble learning not only leads to more accurate models but also provides more robust performance counter ranking. We achieved up to 61.15\% performance improvement and up to 62.58\% energy saving for P1B2, and up to 55.81\% performance improvement and up to 52.60\% energy saving for NT3 on up to 24,576 cores on Cray XC40 Theta.

Overall, ensemble learning  provides a broad, robust view of learning from the data. Applying ensemble learning to science should help better identify key factors that impact science or decision-making [PR06]. We learned from this work that all HPC systems support the number of huge page sizes to accelerate scientific applications with recompiling requirement; however, there is no huge page size support for the ML and deep learning applications. For the deep learning applications and others using script languages such as Python, the huge page support should accelerate the time-consuming training process as well.  Which huge page size will result in the best performance, however,  depends on the application characteristics and the underlying systems.


\section*{Acknowledgments}
This work was supported in part by Laboratory Directed Research and Development (LDRD) funding from Argonne National Laboratory, provided by the Director, Office of Science, of the U.S. Department of Energy under contract DE-AC02-06CH11357; in part by the U.S. Department of Energy, Office of Science, under contract DE-AC02-06CH11357; and in part by NSF grant CCF-1801856. We acknowledge the Argonne Leadership Computing Facility (ALCF) for use of Theta under the DOE INCITE project CANDLE and ALCF project EE-ECP. We acknowledge Huihuo Zheng from ALCF for the TensorFlow and Horovod environments on Theta.

\begin{thebibliography}{}
%
%

\bibitem{BJ12} W. Bircher, and L. John, Complete System Power Estimation Using Processor Performance Events. IEEE Tran. on Comp., 61(4), 2012.
\bibitem{BS13} R. Bertran, Y. Sugawara, H.M. Jacobson, A. Buyuktosunoglu, and P. Bose, Application-Level Power and Performance Characterization and Optimization on IBM Blue Gene/Q systems, IBM J.  Res. and Dev., 57(1/2), 2013.
\bibitem{8} CANDLE: Cancer Distributed Learning Environment,  http://candle.cels.anl.gov.
\bibitem{21} CANDLE Benchmarks: https://github.com/ECP-CANDLE/Benchmarks (Accessed in Sep. 2019).
\bibitem{CX10} X. Chen, C. Xu, R. Dick, and Z. Mao, Performance and Power Modeling in a Multi-Programmed Multi-Core Env., DAC2010, 2010.
\bibitem{CH19} T. Chen, T. He, M. Benesty, et al., Extreme Gradient Boosting, Package ``xgboost,'' August 1, 2019.
\bibitem{CD06} M. Curtis-Maury, J. Dzierwa, C. Antonopoulos, and D. S. Nikolopoulos, Online Power-Performance Adaptation of Multithreaded Programs Using Hardware Event-Based Prediction, ICS'06, 2006.
\bibitem{CS08} M. Curtis-Maury, A. Shah, F. Blagojevic, D. S. Nikolopoulos, B. R. de Supinski, and M. Schulz, Prediction Models for Multidimensional Power-Performance Optimization on Many Cores, PACT?08.
\bibitem{CM05} G. Contreras and M. Martonosi, Power Prediction for Intel XScale Processors Using Performance Monitoring Unit Events, ISLPED?05.
\bibitem{FL12} P. Flach, Machine Learning: The Art and Science of Algorithms that Make Sense of Data, Cambridge University Press, 2012.
\bibitem{FJ01} J. H. Friedman, Greedy function approximation: a gradient boosting machine, Annals of Statistics, 1189--1232, 2001. 
\bibitem{FH10} J. Friedman, T. Hastie, and R. Tibshirani, Regularization Paths for Generalized Linear Models via Coordinate Descent,  Journal of Statistical Software 33(1), 2010.
\bibitem{GF10} R. Ge, X. Feng, S. Song, et al., PowerPack: Energy Profiling and Analysis of High-Performance Systems and Applications, IEEE Trans. on Para. and Dis. Sys. 21(5), 2010.
\bibitem{GLM} glmnet: Lasso and Elastic-Net Regularized Generalized Linear Models, https://cran.r-project.org/web/packages/glmnet/vignettes/glmnet\_beta.pdf.
\bibitem{GL18} J. L. Greathouse and G. H. Loh,  Machine Learning for Performance and Power Modeling of Heterogeneous Systems (Invited Paper), in IEEE/ACM International Conference on Computer-Aided Design (ICCAD'18), San Diego, CA, November 5--8, 2018.
\bibitem{HH20} T. Hothorn, K. Hornik, C. Strobl, and A. Zeileis, A Laboratory for Recursive Partytioning, Package ``Party,'' March 5, 2020.
\bibitem{IM03} C. Isci and M. Martonosi, Runtime Power Monitoring in High-End Processors: Methodology and Empirical Data, 36th IEEE/ACM Intern. Sym. on Microarchitecture, 2003.
\bibitem{KSH} A. Karatzoglou, A. Smola, and K. Hornik, kernlab -- An S4 Package for Kernel Methods in R, Journal of Statistical Software, 11(9), 1--20, 2004.
\bibitem{KF05} N. Kappiah, V. Freeh, and D. Lowenthal. Just In Time Dynamic Voltage Scaling: Exploiting Inter-Node Slack to Save Energy in MPI Programs, SC05, 2005.
\bibitem{KJ13} M. Kuhn and  K. Johnson, Applied Predictive Modeling, Springer, 2013. 
\bibitem{1} M. Kuhn, The caret Package, https://topepo.github.io/caret/index.html, March 27, 2019 (https://cran.r-project.org/web/packages/caret/). 
\bibitem{KW20} M. Kuhn, S. Weston, C. Keefer, N. Coulter, and R. Quinlan, Rule- and Instance-Based regression Modeling, Package``Cubist,'' January 10, 2020. https://topepo.github.io/Cubist. 
\bibitem{LS10}	D. Li, B. de Supinski, M. Schulz, K. Cameron, and D. Nikolopoulos, Hybrid MPI/OpenMP Power-Aware Computing,  IEEE Intern. Conf. on Para. \& Dist. Proc. Symp. (IPDPS2010), 2010.
\bibitem{LW18} A. Liaw, and M. Wiener, Breiman and Cutler's random forests for classification and regression, Package ``randomForest,'' March 25, 2018.
\bibitem{LP10} Lim M, Porterfield A, Fowler R (2010) SoftPower: Fine-Grain Power Estimations Using Performance Counters,  19th Intl. High Perf. Dis. Comp. (HPDC ?10), 2010.
\bibitem{LP13} J. H. Laros III, P. Pokorny, and D. DeBonis, PowerInsight -- A Commodity Power Measurement Capability, International Green Computing Conference, 2013. 
\bibitem{LW11} C. Lively, X. Wu, V. Taylor, S. Moore, H. Chang, and K. Cameron, Energy and Performance Characteristics of Different Parallel Implementations of Scientific Applications on Multicore Systems, Intern. J. High Perf. Comp. and App., 25(3), 2011.
\bibitem{LT14} C. Lively, V. Taylor, X. Wu, H. Chang, C. Su, K. Cameron, S. Moore, and D. Terpstra, E-AMOM: An Energy-Aware Modeling and Optimization Methodology for Scientific Applications on Multicore Systems, Comp. Sci. -- Res. and Dev.,  29(3), 2014. 
\bibitem{LW12} C. Lively, X. Wu, V. Taylor, S. Moore, H. Chang, C. Su, and K. Cameron, Power-Aware Predictive Models of Hybrid (MPI/OpenMP) Scientific Applications on Multicore System, Comp. Sci. -- Res. and Dev., 27(4), 2012.
\bibitem{MP16} P. Miller, mvtboost example, Dec. 5, 2016, https://cran.r-project.org/web/packages/mvtboost/vignettes/mvtboost\_vignette.html
\bibitem{MZ16} Z. Mayer, A Brief Introduction to caretEnsemble, Jan. 31, 2016.
\bibitem{ML02} A. Miyoshi, C. Lefurgy, E. V. Hensbergen, R. Rajamony, and R. Rajkumar, Critical Power Slope: Understanding the Runtime Effects of Frequency Scaling, ICS?02, June 2002.
\bibitem{MS19} S. Milborrow, Multivariate Adaptive Regression Splines, package ``earth,'' November 9, 2019.
\bibitem{MuMI} MuMMI project, http://www.mummi.org/info.
\bibitem{NM10} H. Nagasaka, N. Maruyama, A. Nukada, T. Endo, and S. Matsuoka, Statistical Power Modeling of GPU Kernels Using Performance Counters, Intern. Green Comp. Conf., 2010.
\bibitem{PAPI} PAPI (Performance API), http://icl.cs.utk.edu/papi/
\bibitem{PERF} perf\_event, https://perf.wiki.kernel.org/index.php/Main\_Page. 
\bibitem{PMON} PerfMon2, http://perfmon2.sourceforge.net.
\bibitem{PR06} R. Polikar, Ensemble Based Systems in Decision Making, IEEE Circuits and Systems Magazine, 2006. 
violin package: https://cran.r-project.org/web/packages/vioplot/index.html. https://www.r-graph-gallery.com/violin-plot/
\bibitem{PO08} C. Pospiech, Hardware Performance Monitor (HPM) Toolkit Users Guide, Adv. Comp. Tech. Center, IBM Research, June 2008.
\bibitem{PYPA} PyPAPI, https://flozz.github.io/pypapi
\bibitem{RA13} R. Rodrigues, A. Annamalai, I. Koren, and S. Kundu, A Study on the Use of Performance Counters to Estimate Power in Microprocessors, IEEE Trans. on Cir. and Sys., 60(12), 2013.
\bibitem{RN12} E. Rotem, A. Naveh, D. Rajwan, A. Ananthakrishnan, and E. Weissmann,  Power-management Architecture of the Intel Micro-architecture Code-named Sandy Bridge, IEEE Micro, 32(2), 2012.
\bibitem{2} scikit-learn, Machine Learning in Python, https://scikit-learn.org/stable/index.html
\bibitem{SB08} K. Singh, M. Bhadhauria, and S. A. McKee, Real Time Power Estimation and Thread Scheduling via Performance Counters, Workshop on Design, Arch., and Sim. of Chip Multi-Processors, 2008.
\bibitem{SS13} S. Song, C. Su, B. Rountree and K. Cameron, A Simplified and Accurate Model of Power-Performance Efficiency on Emergent GPU Architectures, IPDPS2013, 2013.
\bibitem{SL06} R. Springer, D. Lowenthal, B. Rountree, and V. Freech, Minimizing Execution Time in MPI Programs on an Energy-Constrained, Power-Scalable Cluster, PPoPP'06, 2006.
\bibitem{SQ08} M. Suleman, M. Qureshi, and Y. Patt, Feedback-Driven Threading: Power-Efficient and High-Performance Execution of Multi-Threaded Workloads on CMPs, ASPLOS?08, 2008. 
\bibitem{SUMM} Summit, https://www.olcf.ornl.gov/olcf-resources/compute-systems/summit/ 
\bibitem{THET} Theta, Cray XC40 system, https://www.alcf.anl.gov/theta.
\bibitem{TL14} G.L. Tsafack Chetsa, L. Lefevre, J.M. Pierson, P. Stolf, and G. Da Costa, Exploiting Performance Counters to Predict and Improve Energy Performance of HPC Systems, Future Generation Computer Systems, Vol. 36, July 2014.
\bibitem{WT14} X. Wu, V. Taylor, C. Lively, H. Chang, B. Li, K. Cameron, D. Terpstra and S. Moore, MuMMI: Multiple Metrics Modeling Infrastructure (Book Chapter), Tools for High Performance Computing 2013, Springer, 2014.
\bibitem{WT16b} X. Wu and V. Taylor, Utilizing Hardware Performance Counters to Model and Optimize the Energy and Performance of Large Scaled Scientific Applications on Power-Aware Supercomputers, in IPDPS2016 Workshop on High Performance, Power-Aware Computing, Chicago, IL, May 23--27, 2016.
\bibitem{WT16} X. Wu, V. Taylor, J. Cook, and P. Mucci, Using Performance-Power Modeling to Improve Energy Efficiency of HPC Applications, IEEE Computer  49(10), 20--29, Oct. 2016.
\bibitem{WT19} X. Wu, V. Taylor, J. M. Wozniak, R. Stevens, T. Brettin, and F. Xia, Performance, Energy, and Scalability Analysis and Improvement of Parallel Cancer Deep Learning CANDLE Benchmarks, in 48th International Conference on Parallel Processing, Kyoto, Japan, August 5--8, 2019.
\bibitem{XLA} XLA architecture, https://www.tensorflow.org/xla/architecture
\bibitem{ZJ17} X. Zheng, L. K. John, and A. Gerstlauer, LACross: Learning-Based Analytical Cross-Platform Performance and Power Prediction, Int. J. Parallel Prog. 45:1488?--1514, 2017. 
\bibitem{ZH18} H. Zou  and T. Hastie, Elastic-Net for Sparse Estimation and Sparse PCA, Package ``elasticnet,'' August 31, 2018.

\end{thebibliography}
\end{document}